\documentclass[journal]{IEEEtran}
\usepackage[english]{babel}
\usepackage[utf8]{inputenc}
\usepackage{algorithm}
\usepackage[noend]{algpseudocode}
\usepackage{makecell}
\usepackage{pbox}
\usepackage[figurename=Fig.]{caption}
\usepackage[font=small,skip=0pt]{caption}
\usepackage{amsmath}
\usepackage{mathptmx}
\usepackage{graphicx}
\usepackage[font=small]{caption}
\usepackage{multirow}
\usepackage{accents}
\usepackage{cite}
\usepackage{comment}
\usepackage{bm}
\usepackage{url}
\usepackage{flushend} 
\usepackage{fancyhdr}

\usepackage[autostyle]{csquotes}

\ifCLASSINFOpdf
\else 
\fi
\begin{document}
\title{A Bayesian Approach with Type-2 Student-\textit{t} Membership Function for T-S Model Identification}

\author{Vikas Singh,
       Homanga Bharadhwaj
        and Nishchal K Verma 
\thanks{Vikas Singh Homanga Bharadhwaj, and Nishchal K Verma  are with the Department of Electrical Engineering, IIT Kanpur, India (e-mail:  vikkyk@iitk.ac.in, homangab@iitk.ac.in, nishchal.iitk@gmail.com)}
}
\markboth{}%
{Shell \MakeLowercase{\textit{et al.}}: Bare Demo of IEEEtran.cls for IEEE Journals}
\maketitle
\begin{abstract}
Clustering techniques have been proved highly successful for Takagi-Sugeno (T-S) fuzzy model identification. In particular,  fuzzy \textit{c}-regression clustering  based on type-2 fuzzy set has been shown the remarkable results on non-sparse data but their performance degraded on sparse data. In this paper, an innovative architecture for  fuzzy \textit{c}-regression model is presented and a novel student-\textit{t} distribution based membership function is designed for sparse data modelling. To avoid the overfitting, we  have adopted a Bayesian approach for incorporating a Gaussian prior on the regression coefficients.  Additional novelty of our approach lies in type-reduction where the final output is computed using Karnik Mendel algorithm and the consequent parameters of the model are optimized using Stochastic Gradient Descent method. As detailed experimentation, the result shows that proposed approach outperforms on standard datasets in comparison of various state-of-the-art methods. 
\end{abstract}
\IEEEpeerreviewmaketitle
\section{Introduction}
\IEEEPARstart{T}{here} have been numerous research focusing on modeling of non-linear systems through their input and output mapping.
In particular, fuzzy logic based approaches have been very successful in modeling of non-linear dynamics in the presence of  uncertainties \cite{ c18}. Type-1 fuzzy logic enables system identification and modeling by virtue of numerous linguistic rules. Although this approach performs well,  but due to the limitation of crisp membership values, its potential to handle the uncertainty in data is limited. Therefore, in order to successfully model the data uncertainty, a type-2 fuzzy logic was proposed, where membership values of each data points are themselves fuzzy. The type-2 fuzzy logic has been remarkably successful in past due to its robustness in the presence of imprecise and noisy data \cite{c101, c102}.

The basic steps used in fuzzy inference system are the structure and parameter identification of the model. The structure identification is related to the process of selecting number of rules,  input features and partition of input-output space while parameter identification is used to compute the  antecedent and consequent parameters of the model. In the  literature's fuzzy clustering have been widely used for fuzzy space partitioning since, a T-S fuzzy model is comprised of various locally weighted linear regression models, many of them are hyperplane based models incorporating a hyperplane based clustering and seems very effective for structure identification. In particular, fuzzy \textit{c}-regression clustering that are hyperplane-shaped clustering becomes more popular \cite{ c31, c32 }. The architecture of these algorithms are robust in partitioning the data space, inferring estimates of outputs with the inputs and  determining an optimum fit for the regression model.  Previously proposed techniques like  fuzzy \textit{c}-regression model (FCRM) and fuzzy \textit{c}-mean (FCM)  have  been developed for type-2 fuzzy logic framework. Here, upper and lower membership values are determined by simultaneously optimizing two objective functions. Interval type-2 (IT2) FCRM, which was presented recently for the T-S regression framework has shown significantly better performance in terms of error minimization and robustness in comparison of type-1 fuzzy logic  \cite{c31, c32,c13}.
			
In this paper, we have combined the Gaussian and student-\textit{t} density type membership function for an IT2 FCRM framework. This is a hyperplane-shaped membership function with  relatively two different weighed terms. The student-\textit{t} density part is weighed more if the data being modeled is sparse. The student-\textit{t} distribution is a popular prior for sparse data modelling in Bayesian inference. Therefore, it is used in our model. The stochastic gradient descent (SGD) technique is used for optimization  of consequent parameters and Karnik Mendel (KM) algorithm is applied for type reduction of estimated output.  We have used $\textit{L}_2$ regularization of the regression coefficients in the IT2 fuzzy \textit{c}-means clustering for identification of antecedent parameters.   As demonstrated in the results section, the regularization helps our model against overfitting of the training data and increases the generalization on unseen data. In addition,  an innovative scheme for optimizing the consequent parameters is also presented, wherein we do not perform type reduction of the type-1 weight sets prior for output estimation. Instead, we use KM algorithm for the output to infer an optimal interval type-1 fuzzy  set and  the set boundaries are optimized by SGD method. 

The rest of the paper is organized as: In Section II, we discuss  the TSK fuzzy model, IT2-FCR and IT2-FCRM. In Section III, we  describe the proposed approach . In Section IV, we present the efficacy of proposed approach through  experimentation.  Finally,  Section V concludes the paper. 
\hfill 
\hfill 
\section{Preliminaries}
\subsection{TSK Fuzzy Model}
TSK fuzzy model provides a rule-based structure for modeling a complex non-linear system.  If $G(x,y)$ is the system to be identified, where $\mathbf{x}\in R^n$ be the input vector and $y \in R $ be the  output. Then, the $i^{th} $ rule is written as

Rule i : IF $x_i$ is $A^i_1$ and $\cdots$ \text{and}$\;$$x_m$ is $A^i_m$ THEN
\begin{align}
y^i = \theta_0^i + \theta_1^ix_1 +\cdots+\theta_m^ix_m 
\end{align}

where, $ i = 1,  \cdots,c $ is the number of fuzzy rule and $y^i$ is the $i^{th}$ output. Using these rules, we can infer the final model output as follows:
\begin{align}
y = \frac{\sum^c_{i=1}w^iy^i}{\sum^c_{i=1}w^i},	\; \;\; w^i = \prod^m_{j=1}\mu_{A_j}^ix_j
\end{align}

 where, $w^i$ denotes the overall firing strength of the $i^{th}$ rule.


\subsection{Interval Type-2 FCM (IT2-FCM)}
In the  interval type-2 FCM, two different objective function that differ in their degrees of fuzziness are optimized simultaneously using Lagrange multipliers to obtain the upper and lower membership function \cite{c102}. Let $m_1$ and $m_2$ be the two degree of fuzziness, then the two objective function are  described as
\begin{align}
\begin{split}
Q_{m_1}(U,v) = \sum_{k=1}^N\sum_{i=1}^c\mu_i(\textbf{x}_{k})^{m_1}{E_{ik}}(\zeta_i)^2\\
Q_{m_2}(U,v) = \sum_{k=1}^N\sum_{i=2}^c\mu_i(\textbf{x}_{k})^{m_1}{E_{ik}}(\zeta_i)^2
\end{split}
\end{align}


\subsection{Inter Type-2 Fuzzy \textit{c}-Regression Algorithm (IT2-FCR)}
The main motivation of inter type-2 fuzzy \textit{c}-regression algorithm is to partition  the set of $n$ data points $(\mathbf{x}_k,y_k)$ $(k = 1\cdots n)$ into \textit{c} clusters. The data points in every cluster $i$ can be described by a regression model as   
\begin{align}
\begin{split}
\hat{y}_k &= g^i(x_k,{\zeta_i}) =  b^i_1x_{k1} + \cdots
+b^i_mx_{km} + b^i_0 = [\mathbf{x}_k \;1] \zeta_i^T
\end{split}
\end{align}

where, $\mathbf{x}_k = [x_{k1}, \cdots, x_{km}]$ be the $k^{th}$ input vector,  $ j =1, \cdots, m$ be the number of features,  $ i= 1, \cdots,c$  be the number of clusters and $ \mathbf{\zeta}_i=[b^i_1, \cdots, b^i_m, b^i_0]$ be the coefficient vector of the $i^{th}$ cluster. In  \cite{c31}, the coefficient vectors are  optimized  by  weighted least square method, whereas, in our approach we used SGD. The primary objective for using SGD is to make the algorithm robust even for the cases where $[\mathbf{x}^T\mathbf{P}_i\mathbf{x}]$   become singular \cite{c13}.
\section{Proposed Methodology}
In this paper we have presented a new framework for FCRM with an innovative student-$t$ distribution  based membership function  (MF) for sparse data modelling \cite{c13,ct, MAP}. The presented approach is described in following subsections.

\subsection{Fuzzy-Space Partitioning}
Firstly, we formulate the task of fuzzy-space partitioning as a Maximum \textit{A-Posterior} (MAP) over a squared error function  \cite{MAP}. Exploiting Bayes rule the MAP estimator is defined as
\begin{align}
\phi(y) = \arg \max_{x\in{R^n}}p(x/y)=\arg \max_{x\in{R^n}}p(y/x)p(x)
\end{align}

where, $p(x/y)$ be the posterior, $p(y/x)$ be the likelihood and $p(x)$ be the prior distribution. Using the  above equation the MAP estimator is expressed in term of regression problem as
\begin{align}
{E_{ik}}(\zeta_i) = (y_k - g^i(x_k,{\zeta_i}))^2 + \lambda\sum_{p=0}^{m}(b^i_p)^2
\end{align}

where,  ${E_{ik}}(\zeta_i)$ be the MAP estimator, $(y_k - g^i(x_k,{\zeta_i}))^2$   be the likelihood,  $\sum_{p=0}^{m}(b^i_p)^2$ be the prior or called  as a regularizer, which is equivalent to the Bayesian notion of having a prior on the regression weights $b^i$ for each cluster $i$ and $\lambda$ be the regularizer control parameter. The regularizer reduces the overfitting in cluster assignment by constraining the small regression weights. 

In the proposed approach, we first define two degrees of fuzziness $m_1$ and $m_2$, initializes the number of clusters $c$ and a termination threshold $ \epsilon $. We also initialize the parameters $\overline{\zeta_i}$ and $\underline{\zeta_i}$, which are the upper and lower regression coefficient vectors of $i^{th}$ cluster. Then, the equation (7) is written in the term upper and lower error function MAP estimator as follows:
\begin{align}
\begin{split}
{E_{ik}}(\overline\zeta_i) = (y_k - g^i(x_k,\overline{\zeta_i}))^2 + \lambda\sum_{p=0}^{m}(b^i_p)^2\\
{E_{ik}}(\underline\zeta_i) = (y_k - g^i(x_k,\underline{\zeta_i}))^2 + \lambda\sum_{p=0}^{m}(b^i_p)^2
\end{split}
\end{align}

To reduce the complexity of the system, a weighted average type reduction technique
is applied  to obtain $ {E_{ik}}(\zeta_i) $ as
\begin{align}
{E_{ik}}(\zeta_i)= \frac{({E_{ik}}(\overline\zeta_i)+{E_{ik}}(\underline\zeta_i))}{2}
\end{align} 

Through MAP estimate on the posterior of defuzzified error function ${E_{ik}}(\zeta_i)$, the upper and lower membership function for every data points in each cluster are obtained as similar to  \cite{c102} and they are given as follows:
\begin{align}\label{eqt2}
\begin{split}
        \overline u_{ik}=\left\{
                \begin{array}{ll}
                  \frac{1}{\sum_{r=1}^{c}\Big (\frac{(E_{ik}(\zeta_i)}{E_{rk}(\zeta_i)}\Big)^\frac{2}{(m_{1}-1)}} ,  &   \text{if}\;\; \frac{1}{\sum_{1}^{c}\Big (\frac{(E_{ik}(\zeta_i)}{E_{rk}(\zeta_i)}\Big)} < \frac{1}{c} \\
                  \frac{1}{\sum_{r=1}^{c}\Big (\frac{(E_{ik}(\zeta_i)}{E_{rk}(\zeta_i)}\Big)^\frac{2}{(m_{2}-1)}}, & \text{otherwise} \\
                \end{array}
              \right. \\            
   \underline u_{ik}=\left\{
                \begin{array}{ll}
                  \frac{1}{\sum_{r=1}^{c}\Big (\frac{(E_{ik}(\zeta_i)}{E_{rk}(\zeta_i)}\Big)^\frac{2}{(m_{1}-1)}} ,  &   \text{if}\;\; \frac{1}{\sum_{1}^{c}\Big (\frac{(E_{ik}(\zeta_i)}{E_{rk}(\zeta_i)}\Big)} \geq \frac{1}{c} \\
                  \frac{1}{\sum_{r=1}^{c}\Big (\frac{(E_{ik}(\zeta_i)}{E_{rk}(\zeta_i)}\Big)^\frac{2}{(m_{2}-1)}}, & \text{otherwise} \\
                \end{array}
              \right. 
 \end{split}
\end{align}

The above equation can be interpreted as that for a MAP  problem formulated in (8). To estimate the parameters $\overline{\zeta_i}$ and $\underline{\zeta_i}$, we formulate the problem as a locally weighted linear regression with an objective function:
\begin{align}
    J(\zeta_i) = \frac{1}{2}\sum_{k=1}^n u_{ik}([\mathbf x_k \;  1]\zeta_i^T - y_k)
\end{align}

Here,  $u_{ik}$ denotes the membership value of  $k^{th}$  data point in the $i^{th}$  cluster. The parameter $\overline{\zeta_i}$ and $\underline{\zeta_i}$ are estimated by SGD using the above objective function by appropriately finding  $\overline u_{ik}$ and $ \underline u_{ik}$.  Then the regression coefficient ($\zeta_i $) are obtained by a type reduction technique as follow:
\begin{align}
\zeta_i = \frac{(\overline\zeta_i + \underline\zeta_i)}{2}
\end{align}

The steps in this subsection are run and the parameters are updated until  the convergence of $||\zeta_i^{current} - \zeta_i^{previous}||$ $\geq$ $\epsilon$ to obtain the optimal value of the regression coefficient as briefly described in Algorithm 1.

\subsection{Identification of Antecedent Parameters}
The MF developed in \cite{c32} is hyperplane shaped, which cannot successfully incorporate the relevant information of data distributions within different clusters. To overcome this issue, we proposed a modified Gaussian based MF combined with a student-\textit{t} density function. The student-\textit{t} distribution is widely used as a prior in Bayesian inference for sparse data modelling  \cite{ct}. Here, we weigh the Gaussian and the student-\textit{t} part by a hyper-parameter $\alpha$. If the data we are modelling is very sparse then, $\alpha$ should be set very low so as to give more weight to student-\textit{t} density membership value.
\begin{align}
\begin{split}
\overline\mu_{A_i}(\mathbf x_k)= \alpha \exp\left(-\eta\frac{(d_{ik}(\overline\zeta_i) - v_i(\overline\zeta_i))^2}{\sigma^2_i(\overline\zeta_i)}\right) \\ + (1 - \alpha) \left(1 + \frac{d_{ik}^2(\overline\zeta_i) }{r} \right)^{-\frac{( r+1)}{2}}
\end{split}
\\
\begin{split}
\underline\mu_{A_i}(\mathbf x_k)= \alpha \exp\left(-\eta\frac{(d_{ik}(\underline\zeta_i) - v_i(\underline\zeta_i))^2}{\sigma^2_i(\underline\zeta_i)}\right) \\ + (1 - \alpha) \left(1 + \frac{d_{ik}^2(\underline\zeta_i) }{r} \right)^{-\frac{(r+1)}{2}}
\end{split}
\end{align}

In the above, $d_{ik}$  is the distance between  $k^{th}$ input vector and $i^{th}$ cluster hyperplane.
\begin{align}
d_{ik}(\overline\zeta_i) =\frac{|\text{x}_k.\overline\zeta_i|}{||\overline\zeta_i||};\;\;\;\;\;\;
d_{ik}(\underline\zeta_i) =\frac{|\text{x}_k.\underline\zeta_i|}{||\underline\zeta_i||}
\end{align}

where,  $r=\max \{ d_{ik}(\overline\zeta_i),\;\; i =1\cdots c\}$ is the maximum distance of  $k^{th}$ input vector from the $i^{th}$ cluster, $v_i$ and $\sigma_i$ denotes the average distance and variance  of each data points from the cluster hyperplane respectively.
\begin{align}
v_i(\zeta_i) = \frac{\displaystyle\sum^n_{k=1}d_{ik}(\zeta_i)}{n};\;\;\;\;\;
\sigma_i(\zeta_i)=\frac{\displaystyle\sum^n_{k=1}(d_{ik}(\zeta_i)-v_i(\zeta_i))^2}{n}
\end{align}

The lower MF ($\underline\mu_{A_i}(\mathbf x_k)$) and upper MF ($\overline\mu_{A_i}(\mathbf x_k)$) are called as weights of the TSK fuzzy model corresponding to $k^{th}$ input belonging to the $i^{th}$ cluster.
\begin{algorithm}[H]
\caption{The Proposed Approach}\label{alg:euclid}
\begin{algorithmic}[1]
\State\textbf{Begin}           
\For{\text{i=1 to m }}   
\State\text{Calculate $\overline{\zeta_i}$ and $\underline{\zeta_i}$ the upper and lower regression } 

 vectors using (10)
\State\text{Calculate errors $E_{ik}(\overline\zeta_i)$, $E_{ik}(\underline\zeta_i) $ using (7)} 
\State\text{Calculate upper  and lower MFs ( $\overline u_{ik}$, $\underline u_{ik}$) using (10)}

\State \textbf{end}
\EndFor

\State\text{The above identifies optimal $\overline{\zeta_i}$ and $\underline{\zeta_i}$ \;\;$\forall \; i \in [1,c]$ }
\For{\text{i=1 to m }}

\State\text{Compute the input MFs using (12) and (13)} 
\State\text{Compute the interval type-2 output $y_k$ using (18)} 
 \State \textbf{end}
\EndFor
\State\textbf{End}
\end{algorithmic}
\end{algorithm}

\subsection{Identification of Consequent Parameters}
In the most of the literature the defuzzification of weights is computed before determining the model output $ \hat{y}_k$. The problem with these approaches are that they do not consider effect of model output which will affect the over all performance of the model. To overcome this problem we evaluated the  $\underline y_k$ and $\overline y_k$  corresponding to the $\underline\mu_{A_i}(\mathbf x_k)$ and $\overline\mu_{A_i}(\mathbf x_k)$ using the KM algorithm  \cite{c101}. The values of $\underline y_k$  and  $\overline y_k$ are optimized parallelly until the convergence. The another advantage of  this approach is that it become more robust in handling noise and provide a confidence interval for every output data points.  
The model output $\underline{y}_k$ and $\overline{y}_k$   corresponding to the weights  $\underline\mu_{A_i}(\mathbf x_k)$ and $\overline\mu_{A_i}(\mathbf x_k)$ are calculate using (1) and (2) as follows: 
\begin{align}
\begin{split}
\underline{y}_k = \frac{\displaystyle\sum^p_{i=1}\overline\mu_{A_i}(\mathbf x_k).(\theta_0^i + \theta_1^ix_{k1}+\cdots+\theta_M^ix_{kM})}{\displaystyle\sum^p_{i=1}\overline\mu_{A_i}(\mathbf x_k)+ \displaystyle\sum^c_{i=p+1}\underline\mu_{A_i}(\mathbf x_k)} \\+\frac{ \displaystyle\sum^c_{i=p+1}\underline\mu_{A_i}(\mathbf x_k).(\theta_0^i + \theta_1^ix_{k1}+\cdots+\theta_M^ix_{kM})}{\displaystyle\sum^p_{i=1}\overline\mu_{A_i}(\mathbf x_k)+ \displaystyle\sum^c_{i=p+1}\underline\mu_{A_i}(\mathbf x_k)}
\end{split}
\\
\begin{split}
\overline{y}_k = \frac{\displaystyle\sum^q_{i=1}\underline\mu_{A_i}(\mathbf x_k).(\theta_0^i + \theta_1^ix_{k1}+\cdots+\theta_M^ix_{kM})}{\displaystyle\sum^q_{i=1}\underline\mu_{A_i}(\mathbf x_k)+ \displaystyle\sum^c_{i=q+1}\overline\mu_{A_i}(\mathbf x_k)} \\+ \frac{ \displaystyle\sum^c_{i=q+1}\overline\mu_{A_i}(\mathbf x_k).(\theta_0^i + \theta_1^ix_{k1}+\cdots+\theta_M^ix_{kM})}{\displaystyle\sum^q_{i=1}\underline\mu_{A_i}(\mathbf x_k)+ \displaystyle\sum^c_{i=q+1}\overline\mu_{A_i}(\mathbf x_k)}
\end{split}
\end{align}

  \begin{table*}
\centering
    \caption{\textsc{\small Comparison of performance on house prices dataset}}
\label{house_r}    
\begin{tabular}{ |c|c|c|c|c|c|c|c| }
\hline
 Model& LR & RR & RBFNN & ITFRCM~\cite{robust} & TIFNN~\cite{robust} & RIT2FC~\cite{robust} & Proposed\\
 \hline
MSE& 0.06 & 0.06 & 0.049 & 0.019& 0.045 & 0.035 & \textbf{0.008}\\
 \hline
Coefficient of Determination& 0.68 & 0.69 & 0.67 & 0.73 & 0.77 & 0.79 & \textbf{0.85}\\
 \hline
  Median Absolute Error& 0.71 & 0.73 & 0.73 & 0.75 & 0.80 & 0.81 & \textbf{0.85}\\
 \hline
\end{tabular}
\begin{flushleft}
\textit{\small LR: Logistic Regression,  RR:  Ridge Regression, RBFNN:  Radial Basis Function Neural Network, ITFRCM: Interval Type-2 Fuzzy c-Means, TIFNN: Type-1 Set-Based Fuzzy Neural Network, RIT2FC: Reinforced Interval Type-2
FCM-Based Fuzzy Classifier}

	\end{flushleft}
\end{table*}



where, $p$ and $q$ are switching points and computed by KM algorithm. We run above mentioned steps  until the convergence of $\overline{y}_k$ and $\underline{y}_k$. Finally , the model output is determined by applying a type reduction technique as
\begin{align}
y_k = \frac{\overline{y}_k + \underline{y}_k}{2}
\end{align}
\section{ Results \& Discussion}

\subsection{House Prices Dataset}
\begin{figure}[h]
\centering
  \includegraphics[width=5.5cm]{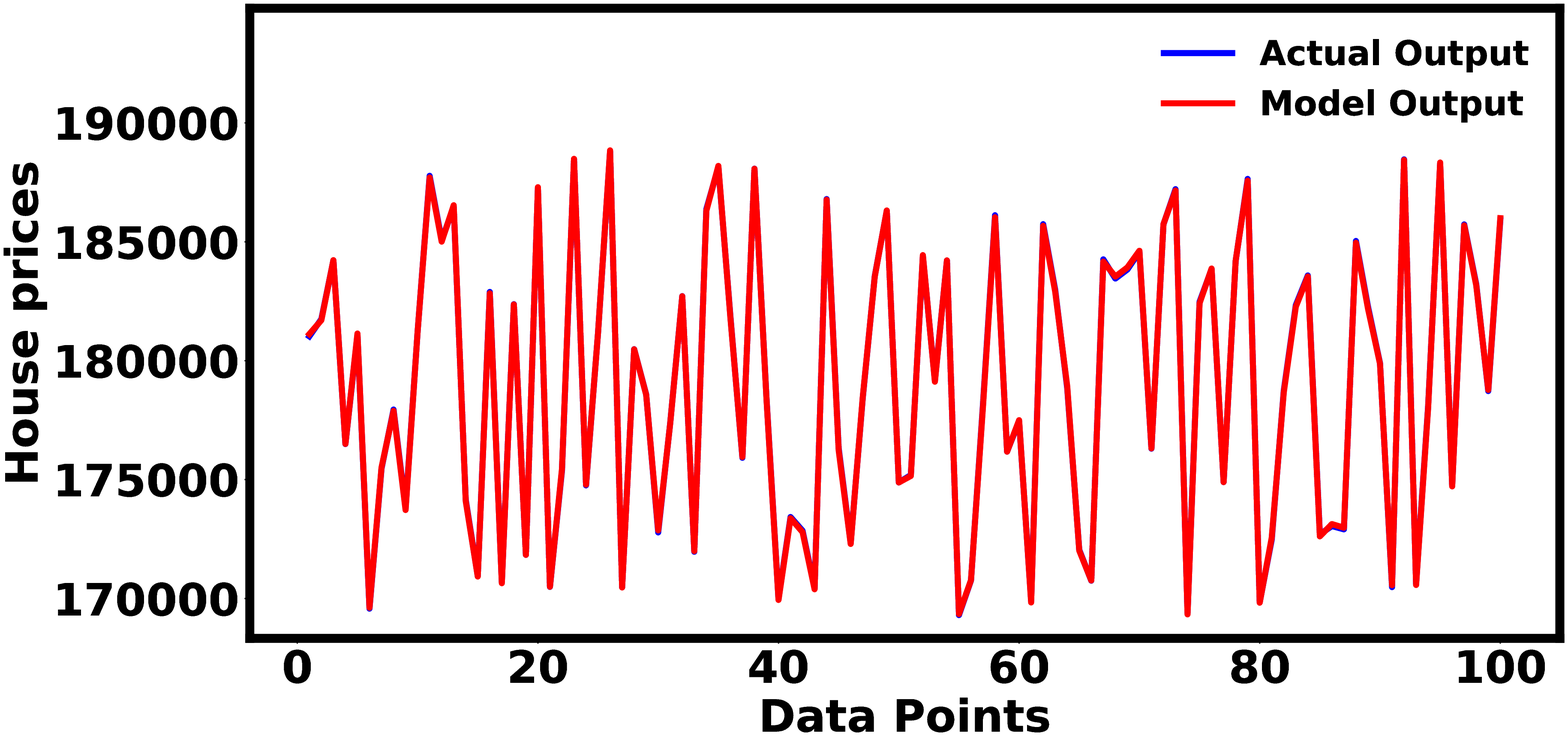}
  \captionof{figure}{Performance comparison of model output with actual output}
  \label{fig:house}
\end{figure}

\begin{figure}[h]
  \centering
  \includegraphics[width=5.5cm]{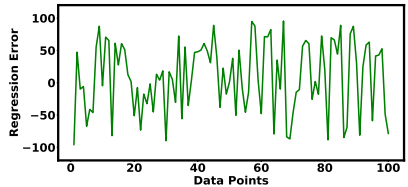}
  \captionof{figure}{Plot of test error for house prices dataset}
  \label{fig:house_error}
\end{figure}

The house prices dataset (\url{https://www.kaggle.com/lespin/house-prices-dataset}) is used to predict the sale price of a particular property.  Through experimentation, we have demonstrated the robustness of proposed method on this sparse data.  The dataset is divided in training (70$\%$) and testing (30$\%$) sets and five-fold cross-validation is used while training. 
The hyper-parameters of the model are  initially set as: $c=3$, $m_1=1.6$, $m_2=4.7$, $\lambda = 0.3$, $\alpha = 0.15$ and $\eta =3.7$.  It should be noted that the value of $\alpha = 0.15$ is small because the dataset is sparse. So, in MF, the contribution of student-$t$  function should be high, which is ensured by a smaller value of $\alpha$ i.e., larger value of $1-\alpha$ as defined by  (12) and (13). The mean square error (MSE)  is $0.008$ on the test data, which is lower than state-of-the-art methods as shown in Table \ref{house_r}. The absolute value of error  as shown in Fig. \ref{fig:house_error} is also small in compare to the absolute house prices as shown in Fig. \ref{fig:house}. We postulate that this is due to the student-$t$  MF  used in our model, which helps in robustly quantifying the effects of sparse data. Also, the higher test accuracy is due to greater generalization owing to $L_2$ regularizer used in our model.   The coefficient of variation which is the ratio of explained variance to total variance is very high ($0.85$).  This  suggests that our model captures variations in the data robustly and is not susceptible to faulty performance in the presence of outliers.

\vspace{-1cm}
\subsection{Non-Linear Plant Modeling }
\begin{figure}[H]
\centering
  \includegraphics[width=5.5cm]{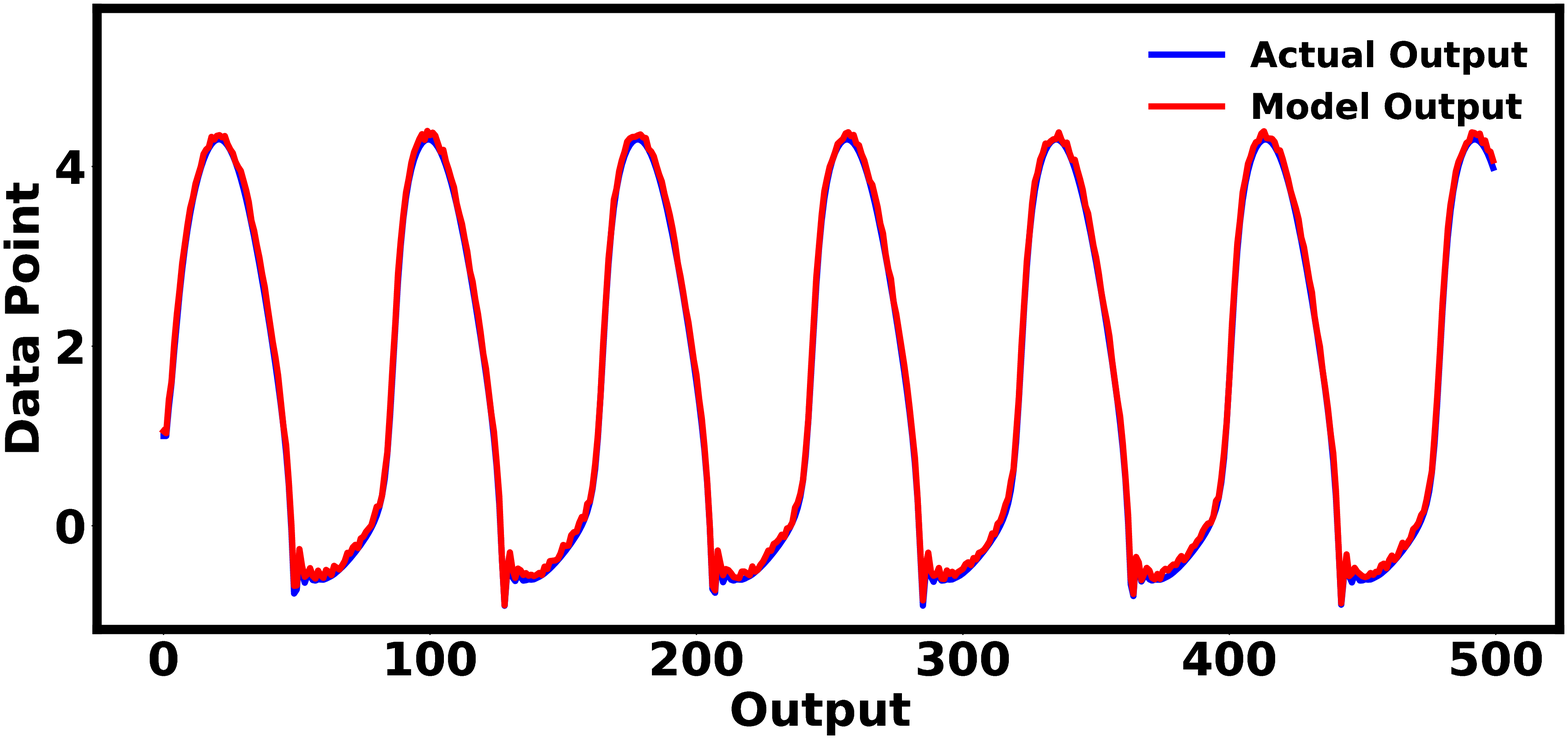}
  \captionof{figure}{Performance comparison of model output and actual output}
  \label{fig:test1}
\end{figure}

\begin{figure}[H]
  \centering
  \includegraphics[width=5.8cm]{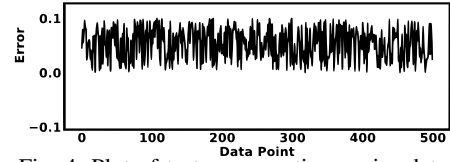}
  \captionof{figure}{Plot of test error on time series data}
  \label{fig:test2}
\end{figure}

The second-order non-linear difference equation as given in (20) is used in order to draw comparison with other benchmark models as given in Table~\ref{nonlinear}.  
\begin{align}
z(k) = \frac{(z(k-1)+2.5)z(k-1)z(k-2)}{1 +z^2(k-1) + z^2(k-2)} + v(k)
\end{align}

 where, $v(k)=\sin(2k/25)$ is the input for validation of model, $z(k)$ is the model output whereas, $z(k-1)$, $z(k-2)$ and $u(k)$ are the model inputs respectively. The hyper-parameters are tuned by grid search and finally set as: $c=4$, $m_1=1.5$, $m_2=7$ and $\eta =3.14$. The obtained MSE of the model on $500$ test data points is $7.2\times 10^{-5}$  using only four rules which is much smaller compared to other models. Through simulations, we have shown that proposed model outperforms with other state-of-the-art model. The Fig. \ref{fig:test1} shows that our model output closely tracks the actual output at every time-step. As observed in Fig. \ref{fig:test2}, the error fluctuates with data point, but the absolute error is consistently less than 0.1 with no rapid surge at stationary points of time series data. This is a crucial requirement for a stable system. Therefore, we conclude that our algorithm yields a dynamically stable model.
\begin{table}[h]
\centering
\setlength\tabcolsep{4pt}
\caption{\textsc{\small Performance on non-linear time series problem}}
\label{nonlinear}
\begin{tabular}{ |c|c|c| }
\hline
 $~~~~~~~~$ State-of-the-Art$~~~~~~~~$ & Rules & MSE\\  
 \hline
   Li et al. \cite{c18} & 4 & $1.49 \times 10^{-2}$     \\ 
 Fazel Zarandi \cite{c31} & 4 & $5.4 \times 10^{-3}$    \\
  Li et al. \cite{c32} & 4 & $1.02 \times 10^{-2}$  \\
 MIT2 FCRM \cite{c13} & 4 & $1.02 \times 10^{-4}$   \\
Proposed  & 4  & $\mathbf{ 7.2\times10^{-5}}$  \\
 \hline
\end{tabular}
\end{table}
\subsection{ A sinc Function in one Dimension}
In this subsection a non-linear \textit{sinc} function is used to present the effectiveness of the proposed model;
\begin{align}
 y = \frac{\sin(x)}{x}
\end{align}

where, $x\in [-40,0) \bigcup (0,40] $. We have sampled 121 data points uniformly for this one dimensional function. As similar to previous case study, the number of rules is taken as four. The hyper-parameters are tuned through grid-search and finally fixed as: $m_1=1.5$, $m_2=7$ and $\eta =3.14$. The  MSE of the proposed model is $2.4\times 10^{-3}$, which is  lower  in compare to  modified inter type-2 FRCM (MIT2-FCRM) \cite{c13}, which is $7.7\times10^{-3}$ on the test data of 121 samples.  The Table~\ref{sine} provides  a detailed comparison of performance with state-of-the-art methods.
\begin{table}[h]
\centering
\setlength\tabcolsep{4pt} 
 \caption{\textsc{\small performance on  \textit{sinc} function}}
\label{sine}
\begin{tabular}{ |c|c|c| }
\hline 
  State-of-the-Art & Rules & MSE\\ 
 \hline
 SCM \cite{chen} & 2 & $4.47 \times 10^{-2}$ \\ 
 EUM \cite{chen} & 2 & $4.50 \times 10^{-2}$ \\ 
 EFCM \cite{chen} & 2 & $8.9 \times 10^{-3}$ \\ 
 Fazel Zarandi  \cite{c31} &  4 & $2.385 \times 10^{-2}$  \\
MIT2 FCRM \cite{c13} & 4 & $7.7 \times 10^{-3}$ \\
 Proposed  &  4 & $\mathbf{2.4 \times 10^{-3}}$ \\
 \hline
\end{tabular}
\end{table}

\section{Conclusion}
In this paper, we have illustrated the efficacy of the proposed  Bayesian type-2 fuzzy regression approach using student-\textit{t} distribution based MF. The proposed MF is useful for  fuzzy $c$-mean regression models as demonstrated in section IV. When the number of features are small in compared to the samples, clustering of  input-output space yield to be very effective for identify the rules of the fuzzy system. 
In addition, we have also demonstrated that instead of direct defuzzification of weights  before computation of the final output, a continuous defuzzification and optimization gives better results.
\ifCLASSOPTIONcaptionsoff
  \newpage			
\fi
 \bibliographystyle{IEEEtran.bst}
\bibliography{IEEEfull.bib}

\end{document}